\documentclass[letterpaper]{article} 
\usepackage{aaai24}  
\usepackage{times}  
\usepackage{helvet}  
\usepackage{courier}  
\usepackage[hyphens]{url}  
\usepackage{graphicx} 
\usepackage{natbib}  
\usepackage{caption} 
\usepackage{algorithm}
\usepackage{algorithmic}

\usepackage{bm}
\usepackage{amsmath}
\usepackage{amsfonts}
\usepackage{diagbox}
\usepackage{multirow}
\usepackage{booktabs}
\usepackage{array}
\usepackage{colortbl}
\usepackage{xcolor}
\usepackage{lipsum}
\usepackage{booktabs}
\usepackage{subcaption}

\usepackage{newfloat}
\usepackage{listings}
\DeclareCaptionStyle{ruled}{labelfont=normalfont,labelsep=colon,strut=off} 
\floatstyle{ruled}
\newfloat{listing}{tb}{lst}{}
\floatname{listing}{Listing}
\title{Multi-Modal Prompt Learning on Blind Image Quality Assessment}
\author {
    Wensheng Pan\textsuperscript{\rm 1}, 
    Timin Gao\textsuperscript{\rm 1},
    Yan Zhang\textsuperscript{\rm 1 \thanks{Corresponding Author.}},
    Runze Hu\textsuperscript{\rm 3},
    Xiawu Zheng\textsuperscript{\rm 1},
    Enwei Zhang\textsuperscript{\rm 2},
    Yuting Gao\textsuperscript{\rm 2},
    Yutao Liu\textsuperscript{\rm 4},
    Yunhang Shen\textsuperscript{\rm 2},
    Ke Li \textsuperscript{\rm 2},
    Shengchuan Zhang\textsuperscript{\rm 1}, 
    Liujuan Cao\textsuperscript{\rm 1}, 
    Rongrong Ji\textsuperscript{\rm 1}
}

\affiliations {
    \textsuperscript{\rm 1}Key Laboratory of Multimedia Trusted Perception and Efficient Computing, Ministry of Education of China, \\ Xiamen University, Xiamen 361005, China\\
    \textsuperscript{\rm 2}Tencent Youtu Lab, Shanghai 200233, China\\
    \textsuperscript{\rm 3}School of Information and Electronics, Beijing Institute of Technology, Beijing 100086, China\\
    \textsuperscript{\rm 4}School of Computer Science and Technology, Ocean University of China, Qingdao 266100, China\\
    bzhy986@xmu.edu.cn
}
\usepackage{bibentry}

\begin{document}
\maketitle

\begin{abstract}
Image Quality Assessment (IQA) models benefit significantly from semantic information, which allows them to treat different types of objects distinctly. Currently, leveraging semantic information to enhance IQA is a crucial research direction.
Traditional methods, hindered by a lack of sufficiently annotated data, have employed the CLIP image-text pretraining model as their backbone to gain semantic awareness. However, the generalist nature of these pre-trained Vision-Language (VL) models often renders them suboptimal for IQA-specific tasks. 
Recent approaches have attempted to address this mismatch using prompt technology, but these solutions have shortcomings. 
Existing prompt-based VL models overly focus on incremental semantic information from text, neglecting the rich insights available from visual data analysis. This imbalance limits their performance improvements in IQA tasks.
This paper introduces an innovative multi-modal prompt-based methodology for IQA.
Our approach employs carefully crafted prompts that synergistically mine incremental semantic information from both visual and linguistic data.
Specifically, in the visual branch, we introduce a multi-layer prompt structure to enhance the VL model's adaptability. In the text branch, we deploy a dual-prompt scheme that steers the model to recognize and differentiate between scene category and distortion type, thereby refining the model's capacity to assess image quality.
Our experimental findings underscore the effectiveness of our method over existing Blind Image Quality Assessment (BIQA) approaches. Notably, it demonstrates competitive performance across various datasets. Our method achieves Spearman Rank Correlation Coefficient (SRCC) values of 0.961(surpassing 0.946 in CSIQ) and 0.941 (exceeding 0.930 in KADID), illustrating its robustness and accuracy in diverse contexts. Our code is available at https://github.com/stephencurry-web/IQA.
\end{abstract}

\begin{figure}[t]
  \centering
    \includegraphics[width=0.48\textwidth]{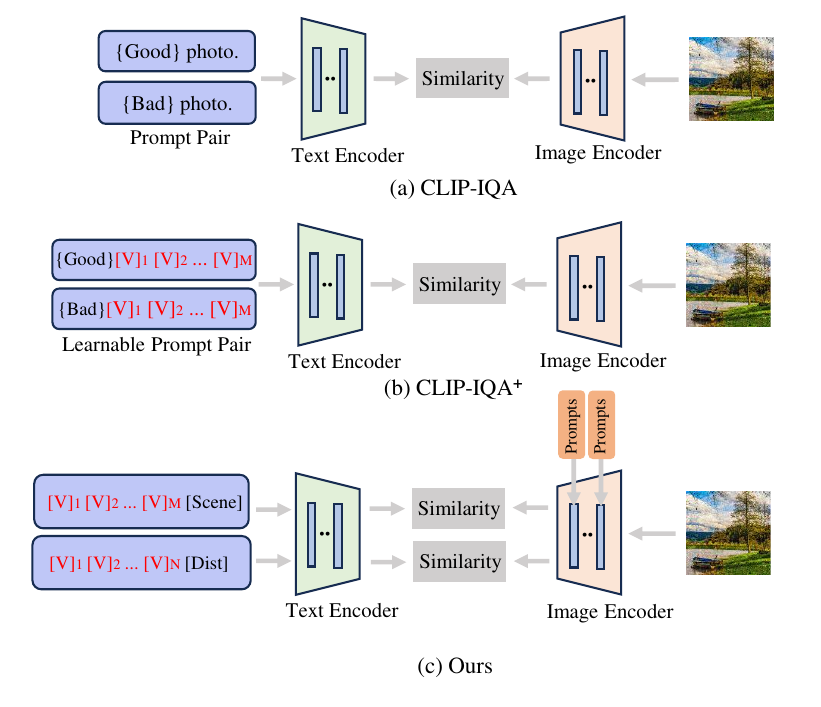}
  
  \caption{Comparison between CLIP-IQA, CLIP-IQA$^{+}$ and the proposed method. (a) CLIP-IQA with antonym prompt pairing strategy. (b) CLIP-IQA$^{+}$ introduced CoOp to learning suitable prompt pairs. (c) Our approach is with (1) a dual-prompt scheme in the text branch and (2) deep prompts in the visual branch.} 
  \label{fig1}
\end{figure}

\section{Introduction}

The aim of Image Quality Assessment (IQA) is to measure and forecast the visual quality of images, taking into account the characteristics of human visual perception.
Due to the unavailability of reference images in most application scenarios, recent research has primarily focused on the Blind Image Quality Assessment (BIQA) domain.
Semantic information greatly enhances BIQA models, enabling them to distinguish between different types of objects.
At present, utilizing semantic information to improve BIQA is a key area of research.

However, traditional methods have resorted to using the CLIP model~\cite{radford2021learning} as a backbone to achieve semantic awareness, due to the scarcity of annotated data in the IQA field.	
For the first time, CLIP-IQA\cite{wang2023exploring} investigated the potential use of the CLIP model in BIQA.
As depicted in Fig.\ref{fig1}-(a), they introduced a novel strategy involving paired prompts with antonymous meanings (e.g., ``Good photo'' and ``Bad photo''). Their experimental results indicated that CLIP can be directly applied to visual perception assessment without the need for fine-tuning for specific tasks. Notably, CLIP demonstrates the ability to discern intricate image features and possess an understanding of abstract attributes such as ``happiness'' and ``sadness''.

While the CLIP model is pre-trained on large-scale text-image pairs to learn a shared semantic space, IQA-specific tasks require a more fine-grained understanding and discrimination ability for IQA. Generic VL models may not provide sufficient details and domain-specific knowledge for IQA tasks, leading to suboptimal performance. Therefore, recent approaches have attempted to introduce prompt learning to guide pre-trained CLIP models toward the downstream task of BIQA. Prompt learning provides a way to guide the model's attention and encourage it to focus on specific aspects of the task. CLIP-IQA$^{+}$~\cite{wang2023exploring} introduced CoOp~\cite{zhou2022coop} to learning suitable prompt pairs, i.e., ``{good} ${[V]}_1 {[V]}_2 ... {[V]}_M$'' and ``{bad} ${[V]}_1 {[V]}_2 ... {[V]}_M$'', as shown in Fig.\ref{fig1}-(b). 

Current prompt-based Visual Language (VL) models in BIQA employ relatively simplistic text prompt designs, which do not adequately capture the nuanced nature of IQA tasks. Image Quality Assessment is inherently subjective and deeply rooted in human visual perception, encompassing a multitude of factors including but not limited to clarity, color fidelity, noise levels, and contrast.
QPT~\cite{zhao2023quality} posit a strong correlation between image quality and the scene category of the image, as well as the type of distortion, and have attempted to capture the differences between various images in these two aspects using contrastive learning. Inspired by this, we have redesigned the text prompts and proposed a dual-prompt scheme, introducing learnable scene category prompts and distortion-type prompts to guide the model to focus on quality-related information in the image.
Specifically, the scene category prompt aims to describe the main content and environment of the image. This prompt is designed to capture the global information of the image, thereby aiding the model in understanding the context of the image. Subsequently, distortion-type prompts focus on describing the quality information of the image, such as blur, noise, or color distortion. This prompt is designed to capture the local information of the image, thereby assisting the model in identifying and evaluating the quality of the image. Two types of prompts work in synergy to assist the model in extracting quality-related image features.

Furthermore, existing approaches primarily focus on incremental semantic information from text, overlooking the valuable insights from visual data analysis. This imbalance restricts performance enhancements in IQA tasks. Inspired by MaPLe's work~\cite{khattak2023maple}, we propose an innovative multi-modal prompt-based methodology for BIQA, integrating prompt learning into both the text and visual branches of CLIP. To improve the VL model's adaptability and broaden the prompt's usefulness, we incorporate a set of learnable vectors into the input of each transformer layer in the image encoder. This design aims to fine-tune the upstream VL model for customized IQA applications. With this multi-modal prompt design, our approach effectively aligns text and visual features, harnesses CLIP's powerful capabilities, and overcomes performance limitations.

Our contributions are the following:
    \begin{itemize}   
        \item To better adapt CLIP to BIQA tasks and leverage its learning ability on text and image, we introduce multi-modal prompt learning for enhanced understanding of visual features and image quality, overcoming CLIP-IQA's performance bottleneck in BIQA tasks. 

        \item In contrast to previous approaches, we propose a novel dual-prompt scheme in the text branch to semantically discern scene categories and distortion types, enhancing the model's grasp of IQA tasks. In the visual branch, a structured visual prompting strategy is employed to progressively enhance the model's evaluative capacity in assessing image quality.
        This integrated approach significantly improves the accuracy of image quality assessment by deeply synergizing textual and visual information.
        
        \item The experimental results show that MP-IQE surpasses state-of-the-art methods on various datasets, offering a novel approach for applying CLIP in BIQA tasks.
        
    \end{itemize}
\section{Related Work}
\subsection{BIQA}
Conventional BIQA research can be further divided into
Natural Scene Statistics (NSS) based~\cite{moorthy2011blind,saad2012blind} or Human Vision System (HVS) based method~\cite{zhai2011psychovisual,gu2014using}. These methods could perform well on synthetically distorted images. However, the images found in real-world scenarios exhibited significant complexities in terms of both distortions and image contents. Conventional methods cannot effectively model real-world distortions.

In recent years, deep learning-based approaches have gradually become mainstream, fostering the development of the BIQA field. The paradigm of these methods~\cite{hypernet, zhang2018blind, TIQA, ke2021musiq, pan2022dacnn, DEIQT, TReS, hypernet} typically involved employing sophisticated neural networks as feature extractors to capture perceptual quality-aware features from images. This was followed by a fully connected layer for mapping the features to quality scores. These methods could be categorized into multi-level feature aggregation~\cite{pan2022dacnn, TReS}, adaptive convolution~\cite{hypernet}, and self-attention~\cite{TIQA, DEIQT, ke2021musiq}.
Adaptive convolution was proposed by HyperIQA~\cite{hypernet}. In this approach, the BIQA process was divided into three stages, including content understanding, perceptual rule learning, and quality prediction. After extracting semantic features from images, a hyper-network was utilized for adaptive rule generation, and these rules were then used in the quality prediction network.
Another significant development in BIQA was the application of self-attention, where the Vision Transformer (VIT)~\cite{dosovitskiy2020vit} was utilized for image quality assessment.
More recently, there have been several emerging paradigms in BIQA. Notably, these included meta-learning for fast adaptation~\cite{zhu2020metaiqa}, unified networks for both synthetic and natural distortions, and continual learning for streaming distortion data~\cite{liu2022liqa, ma2021remember}.

\subsection{CLIP Applications}
CLIP was a multi-modal vision and language model pre-trained on a large corpus of text and image data using a contrastive learning objective. This pretraining process enabled CLIP to learn meaningful representations for images and text, which could then be fine-tuned on specific downstream tasks, such as image classification, and object detection. CLIP has shown impressive performance across various vision and language tasks.

CLIP-IQA first attempted to investigate the potential of CLIP on the challenging yet meaningful task of perception assessment and demonstrated that CLIP could be directly applied to visual perception assessment without task-specific fine-tuning. They introduced a prompt pairing strategy where antonym prompts were adopted in pairs (e.g., “Good photo.” and “Bad photo.”). 
LIQE~\cite{zhang2023mutitask} proposed a paradigm of multi-task learning, which combined BIQA tasks with Distortion Type Identification and Scene Classification to improve the performance of BIQA. Moreover, they employed a Likert scale with five quality levels, and the model output scores as logit weighted sums of the quality levels.

\begin{figure*}[t]
  \centering
   {\includegraphics[width=0.8\linewidth]{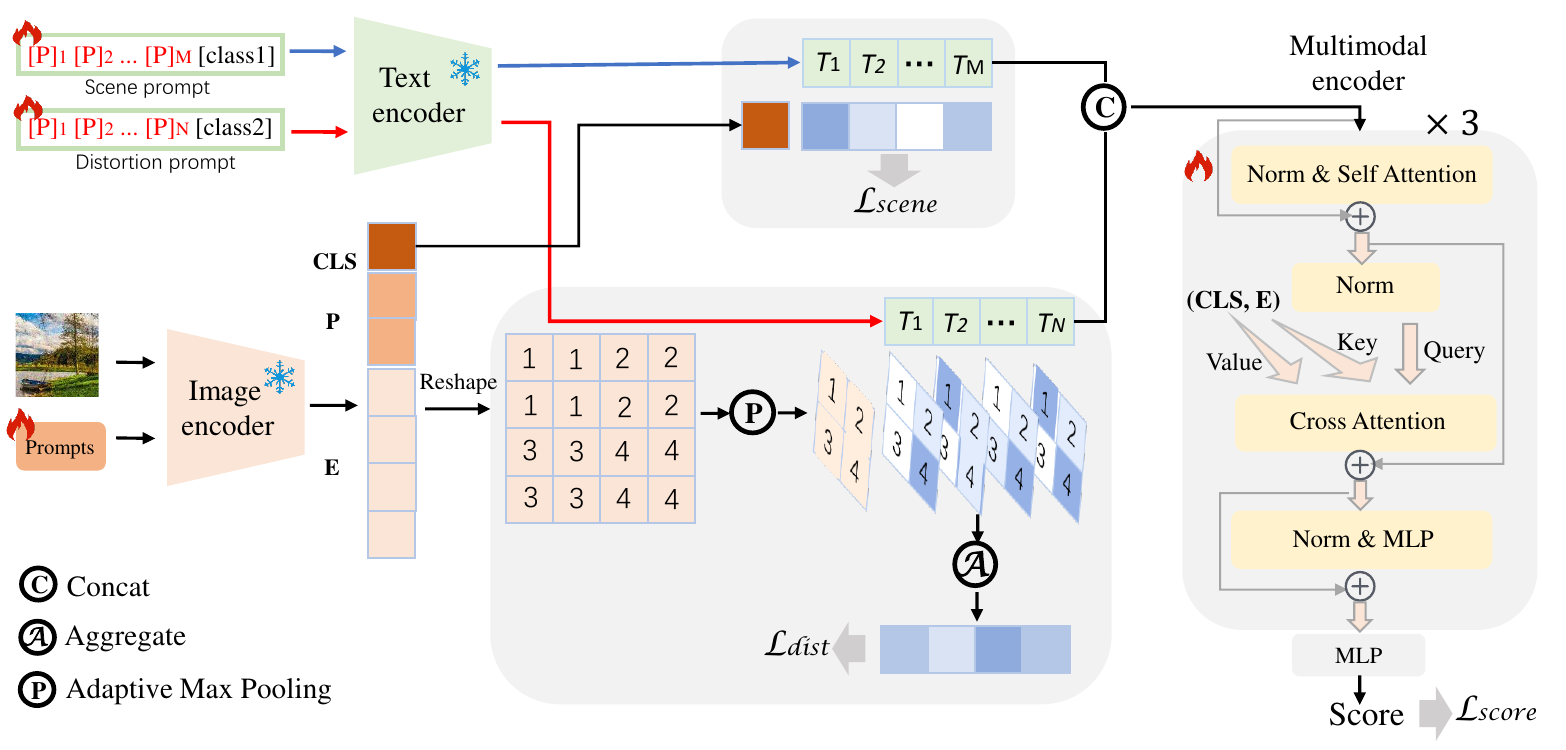}}
  \caption{Overview of our proposed MP-IQE. In the text branch, we introduce the scene prompt and distortion prompt to align the class token embedding and patch embeddings, respectively. In the visual branch, we incorporate a multi-layer prompt to enhance the pre-trained model adaptability. Finally, scene embeddings and distortion embeddings are concatenated as the query of the multi-modal encoder and perform the cross-attention with image features. Then, the multi-modal features are sent to an MLP layer to predict the quality score.}
  \label{fig2}
\end{figure*}

\subsection{Prompt Learning in Vision Language Models}
Inspired by prompt learning in NLP, many works have proposed to adapt V-L models by learning the prompt tokens in end-to-end training.
CoOp introduced learnable prompt vectors to replace fixed text prefixes and only fine-tuned the prompt while keeping other parameters fixed. This approach marked a pioneering achievement by showcasing that prompt learning-based techniques outperform both manually devised prompts and linear probing methodologies. Subsequently, Co-CoOp~\cite{zhou2022cocoop} proposed Conditional Context Optimization to address CoOp's poor generalization on unseen classes. In contrast to the static prompt in CoOp, Co-CoOp uses an instance-adaptive dynamic prompt, rendering it more resilient against class shifts.
Bahng et al.~\cite{bahng2022visual} performed visual prompt tuning on CLIP by prompting on the vision branch. In contrast, Maple was the first to explore multi-modal prompt learning in CLIP to favorably align its vision-language representations. Inspired by their work, we also utilize multi-modal prompt learning to align our image perception features with quality-related text features.


\section{Proposed Method}
\subsection{Preliminaries}

The goal of the BIQA model is to simulate human image quality perception and predict the quality score $q\left(x\right)$ for a given image $x$. 
Like LIQE, we consider nine scene categories: $s \in S $ = \{``animal”, ``cityscape”, ``human”, ``indoor scene”, ``landscape”, ``night scene”, ``plant”, ``still-life”, and ``others”\}.
And identify the dominant distortion in the image: $d \in D$ = \{ ``blur”, ``color-related”, ``contrast”, ``JPEG compression”, ``JPEG2000 compression”, ``noise”, ``over-exposure”, ``quantization”, ``under-exposure”, ``spatially-localized”, and ``others”\} with eleven in total.

\subsection{Overview}

Our goal is to adapt the pre-trained CLIP model using prompt learning for better alignment with the downstream BIQA task. Fig. \ref{fig2} illustrates the overall architecture of our innovative MP-IQE (Multi-Modal Prompt Image Quality Evaluator) framework. 
Compared to the previous fine-tuning CLIP method, we introduce a dual-prompt scheme in the text branch, representing the scene category and distortion type of the image, respectively. 
Meanwhile, we integrate visual prompts at each layer of the image encoder, along with the text prompt, to achieve a more effective alignment between text and image embeddings. 
Moreover, we introduce a multimodal encoder for the interaction between text and image features, using text prompts to guide the model in accurately extracting quality-related semantic information.
In the training stage, we exclusively fine-tune text and visual prompts, and multimodal encoder, keeping other network parameters frozen. 

\subsection{Review CLIP}  
We build our approach on CLIP, which consists of both a text encoder and a vision encoder.

\subsubsection{Text Encoding}
The text encoder employs a multi-layer transformer architecture to encode text descriptions. Each word in a given description (e.g., ``A photo of a [class]") is first converted into a lowercase byte pair encoding (BPE) representation. The BPE representations are then projected to word embeddings $T_0=\left(T_0^1, T_0^2, ..., T_0^N\right),$ where $T_i$ serves as the input for the $i+1$ layer transformer $L_{i+1}$ of the text encoder.
\begin{equation}
    T_{i+1} = L_{i+1} \left( T_i \right)
\end{equation}
Each text sequence is designed to incorporate start [SOS] and end [EOS] tokens.
After passing through the last transformer layers, the final feature representation of the encoded text is obtained by projecting the [EOS] token to the multi-modal embedding space via $Proj_T$.
\begin{equation}
    Z^T = Proj_T \left( EOS \right)
\end{equation}

\subsubsection{Image Encoding}

The initial image embeddings are denoted by $I_0=\left\{CLS, I_0^1, I_0^2, ..., I_0^b\right\}$, where the [CLS] token represents global semantic features and $b$ is the number of patches. Similar to text encoding, after passing through the last layers of transformers, the final image embedding representation $Z^I$ is obtained by projecting the [CLS] token to the multi-modal embedding space via $Proj_I$.

\begin{equation}
    Z^I = Proj_I \left( CLS \right)
\end{equation}
\subsubsection{Align Text and Image}
Let $D=\left\{ x_1, x_2, ..., x_B \right\} $ represent a minibatch of images. The image-to-text contrastive loss ${\mathcal L}_{v2t}$ is calculated as:
\begin{equation}
{\mathcal L}_{v2t}\left(i\right) = -\log \frac{exp\left(Sim\left( Z^T_i,Z^I_i \right)/\mathcal{T}\right)}{\sum_{j=1}^{B} exp\left(Sim\left( Z^T_j,Z^I_i \right)/\mathcal{T}\right)}
\end{equation}
where Sim(·) denotes the similarity score between the text embedding representation and the image embedding representation, $\mathcal{T}$ denotes the temperature that scales the similarity scores. And the text-to-image contrastive loss ${\mathcal L}_{t2v}$:
\begin{equation}
{\mathcal L}_{t2v}\left(i\right) = -\log \frac{exp\left(Sim\left( Z^T_i,Z^I_i \right)/\mathcal{T}\right)}{\sum_{j=1}^{B} exp\left(Sim\left( Z^T_i,Z^I_j \right)/\mathcal{T}\right)}
\end{equation}

\subsection{Multi-Modal Prompt Image Quality Evaluator}

\subsubsection{Scene Prompts and Distortion Prompts}

We enhance the text branch of CLIP by incorporating two types of prompts: scene prompts $\left( P_s \right)$, which capture the overarching content and setting of an image, and distortion prompts $\left( P_d \right)$, which identify specific quality issues such as blur or noise. These prompts are designed to extract distinct semantic information from the image. Taking inspiration from the CoOp framework, which utilizes learnable vectors to adapt prompts to specific tasks, we apply a similar strategy to dynamically model the context within which a prompt operates. This is achieved by integrating a set of learned vectors that can be fine-tuned to optimize the prompt's effectiveness in extracting relevant semantic information. Specifically, the text prompt of the text encoder is designed as follows: 
\begin{equation}
  P = {[V]}_1 {[V]}_2 ... {[V]}_M [class]
\end{equation}
where each ${[V]}_m$ ($m \in \{1, 2, ..., M\}$) is a vector with the same dimension as the text embedding.

\subsubsection{Deep Visual Prompting}

Compared to the previous method, our introduction of Deep Visual Prompting in the vision branch of CLIP involves adding prompts at each layer of the image encoder. 
The main objective is to improve the alignment between the perceptual features of the image and the text features at the quality level.
By integrating prompts at multiple layers, we achieve more precise control over the learning process of image features, leading to improved adaptability to IQA tasks.
Specifically, in our approach, we introduce a set of learnable tokens $\Bar{P} \in \mathbb{R}^{n \times dim}$, which are positioned between the class token and the patch embeddings of the image. 
At each layer of the image encoder, the input is represented as $\left[CLS_{i-1}, \Bar{P}_{i-1}, E_{i-1}\right]$, $E$ denotes the patch embedding. After passing through the ith transformer layer $L_i$, we continue to introduce new learnable tokens ${P}^{\prime}$, which are concatenated with the output $CLS_i$ and $E_i$.
\begin{equation}
    \left[CLS_{i}, \bigcirc, E_{i}\right] = L_i \left( \left[CLS_{i-1},{P}^{\prime}_{i-1}, E_{i-1}\right] \right)
\end{equation}
where $\bigcirc$ means not as input to the next transformer layer. 
This design enables us to incorporate learnable tokens at each layer of the image encoder, thereby enhancing the model's ability to capture and represent essential image features and quality characteristics, ultimately leading to a more effective and accurate image quality assessment.

\begin{table*}[t]
\caption{Performance comparison is measured by averages of SRCC and PLCC. The best results are highlighted in bold, second-best is \underline{underlined}. The results marked with $\ast$ are reproduced based on the experimental settings of the original paper.} 
\setlength\tabcolsep{2pt}
\small
  \centering
  
    \begin{tabular}{lccccccccccccc}
    \toprule
     & \multicolumn{2}{c}{LIVE} & \multicolumn{2}{c}{CSIQ}  & \multicolumn{2}{c}{KADID} & \multicolumn{2}{c}{LIVEC} & \multicolumn{2}{c}{BID} & \multicolumn{2}{c}{KonIQ}\\
    \cmidrule{2-13}
    Method & \multicolumn{1}{c}{PLCC} & \multicolumn{1}{c}{SRCC} & \multicolumn{1}{c}{PLCC} & \multicolumn{1}{c}{SRCC} & \multicolumn{1}{c}{PLCC} & \multicolumn{1}{c}{SRCC}& \multicolumn{1}{c}{PLCC} & \multicolumn{1}{c}{SRCC}& \multicolumn{1}{c}{PLCC} & \multicolumn{1}{c}{SRCC}& \multicolumn{1}{c}{PLCC} & \multicolumn{1}{c}{SRCC} \\
    \midrule
    ILNIQE ~\cite{ILNIQE} & 0.906 & 0.902 & 0.865 & 0.822  & 0.558 & 0.534 & 0.508 & 0.508 & 0.494 & 0.548  & 0.537 & 0.523 
    \\
    UNIQUE ~\cite{zhang2021uncertainty} & 0.952 & 0.961 & 0.912 & 0.902  & 0.875 & 0.852 & 0.884 & 0.854 &0.885 & \textbf{0.884} & 0.900 & 0.895 
    \\
    DBCNN ~\cite{zhang2018blind} & 0.971 & 0.968 & \underline{0.959} & \underline{0.946}  & 0.856 & 0.851 & 0.869 & 0.851 & 0.883 & 0.864  & 0.884 & 0.875  
    \\
    MetaIQA ~\cite{zhu2020metaiqa} & 0.959 & 0.960  & 0.908 & 0.899& 0.775 & 0.762  & 0.835 & 0.802 & 0.868 & 0.856  & 0.887 & 0.850  
    \\
    HyperIQA ~\cite{hypernet} & 0.966 & 0.962 & 0.942 & 0.923 & 0.845 & 0.852 & 0.882 & 0.859 & 0.868 & 0.848   & \underline{0.917} & 0.906  
    \\
    TReS  ~\cite{TReS} & 0.968 & 0.969 & 0.942 & 0.922 & 0.858 & 0.859 & 0.877 & 0.846 & 0.871 & 0.853  & \textbf{0.928} & 0.915 
    \\
    MUSIQ ~\cite{ke2021musiq} & 0.911 & 0.940  & 0.893 & 0.871  & 0.872 & 0.875 & 0.746 & 0.702 & 0.774 & 0.744 & \textbf{0.928} & \underline{0.916}
    \\
    DACNN ~\cite{pan2022dacnn} & \textbf{0.980}  & \textbf{0.978} & 0.957 & 0.943 & 0.905 & 0.905 & 0.884 & 0.866 & $0.759^{\ast}$ & $0.754^{\ast}$ & 0.912 & 0.901  \\

    LIQE ~\cite{zhang2023multitask_clip} & 0.951  & 0.970 & 0.939 & 0.936 & \underline{0.931} & \underline{0.930} & \underline{0.910} & \textbf{0.904} & \textbf{0.900} & \underline{0.875}  & 0.908 & \textbf{0.919}
    \\
    
    \midrule
        MP-IQE(ours) &\textbf{0.980} & \textbf{0.978} & \textbf{0.968} & \textbf{0.961}& \textbf{0.944} & \textbf{0.941}  & \textbf{0.916} & \underline{0.900} & \underline{0.887} & 0.862 & 0.904 & 0.898 
    \\
    
    \bottomrule
    \end{tabular}
  
  \label{table1}
\end{table*}

\subsubsection{Feature Alignment}

After passing through the text encoder, we obtain the text feature $V^s \in \mathbb{R}^{M \times dim}$ representing the scene category and the text feature $V^d \in \mathbb{R}^{N \times dim}$  representing the distortion type. After passing through the image encoder, we obtain the image feature represented by $\left[CLS, P^{\prime}, E\right]$. Given that the original pre-training objective of the CLIP model is to match images with their corresponding text descriptions, which aligns closely with the task of scene recognition, and the scene recognition task requires the representation of global features. Therefore, we align the text features $V^s$ with the $CLS$ features. Calculate the probability that the image belongs to each scene category as follows:

\begin{equation}
p(y=c|x) = \frac{exp\left(Sim\left(CLS,V^s_c \right)/\mathcal{T}\right)}{\sum_{j=1}^{M} exp\left(Sim\left( CLS,V^s_j \right)/\mathcal{T}\right)}
\end{equation}

Then, we use cross-entropy loss to achieve alignment. Represent as:
\begin{equation}
{\mathcal L}_{scene} = - \frac{1}{M} {\sum}^M_{c=1} y_c \log p_c
\end{equation}


The distortion category task requires detailed local information, and as a representation of global information, $CLS$ features cannot fully express the distortion situation in various regions of the image. Therefore, we align the text feature $V^d$ with the patch feature $E \in \mathbb{R}^{b \times dim}$. Specifically, we divide the image into ${w \times w }$ windows, reshape the $E$ into square features, and then perform an adaptive max pooling to obtain the features of each window region. Then calculate similarity with text feature $V^d$ separately. The information from different windows is aggregated in a spatially weighted manner, represented as:

\begin{equation}
p(y=c|x) = {\sum}^{w \times w}_{i=1} \frac{exp\left(Sim\left(P_i,V^s_c \right)/\bar{\mathcal{T}}\right)}{\sum_{i=1}^{w \times w} exp\left(Sim\left(P_i,V^s_c\right)/\bar{\mathcal{T}}\right)} \cdot {Sim\left(P_i,V^s_c \right)}
\end{equation}

Where $\bar{\mathcal{T}}$ denotes the temperature that scales the similarity scores. Then, we apply a softmax operation to the similarity and use the cross entropy function ${\mathcal L}_{dist}$ to align the local features of the image and the text features about the distortion type.
\begin{equation}
{\mathcal L}_{dist} = - \frac{1}{N} {\sum}^N_{c=1} y_c \log p_c
\end{equation}



\subsubsection{Multi-modal Encoder}
In this section, we concatenate the encoded scene prompt and distortion-type prompt to obtain $Q$.
\begin{equation}
Q = Concat\left(V_s, V_d\right)
\end{equation}

 $Q$ is sent to the multi-head self-attention (MHSA) block to obtain $Q^{\prime}$. We use $Q^{\prime}$ as the query for the multi-modal encoder and image features $I^{\prime}=[CLS, E]$ as keys and values. $Q^{\prime}$ and $I^{\prime}$ are sent to a multi-head cross-attention (MHCA) block to perform the cross-attention. During this process, we utilize $Q^{\prime}$
to interact with the features in the encoder outputs, thus ensuring the attentional features are more significant to the image quality. Then, an MLP head derives the final predicted quality score.

\begin{equation}
    Q^{\prime} = MHSA\left(Norm\left(Q\right)\right)+Q, 
\end{equation}

\begin{equation}
    S = MLP \left(MHCA\left(Norm\left(Q^{\prime}\right), I^{\prime},I^{\prime}\right) + Q^{\prime}\right), 
\end{equation}
$Norm\left( \cdot \right)$ indicates the layer normalization.
Through the interaction between text and image, we obtain semantic features closely related to image quality from the image.

\subsubsection{Optimization}
In conclusion, the overall objective function of the MP-IQE can be
formulated as follows:

\begin{equation}
    \mathcal{L}_{total} = {\mathcal L}_{scene} + {\lambda_1} {\mathcal L}_{dist} +  {\lambda_2} {\mathcal L}_{score}
\end{equation}

The hyperparameters $\lambda_1$ and $\lambda_2$ are used to balance different losses and achieve improved performance. The function ${\mathcal L}_{score}$ compute the smooth l1 loss between the predicted score $Y_{pred}$ and the ground truth $GT$, represented as:

\begin{equation}
    {\mathcal L}_{score} = {||Y_{pred} - GT||}_1
\end{equation}


\section{Experiments}
\subsection{Datasets and Evaluation Protocols}
We perform an extensive evaluation of our proposed model using six widely recognized IQA datasets. The datasets we use are LIVE ~\cite{sheikh2006live}, CSIQ ~\cite{larson2010csiq}, KADID ~\cite{lin2019kadid}, BID ~\cite{ciancio2010no}, LIVEC~\cite{ghadiyaram2015livec}, and KonIQ~\cite{hosu2020koniq}. 
The LIVE and CSIQ datasets contain 779 and 866 distorted images, respectively. These images are affected by distortion types, mainly including JPEG2000, JPEG, white noise, and Gaussian blur. KADID comprises 10125 distorted images, utilizing 25 distinct distortion techniques. The LIVEC dataset consists of 1162 images affected by diverse and randomly occurring disruptions, along with genesis capture anomalies commonly found in modern mobile camera devices. 
BID comprises 585 images captured under various scenes, apertures, and exposure times with an 8-megapixel Canon Powershot S3 IS camera, categorized into five blur types ranging from unblurred to complex motion.
Lastly, the KonIQ-10k dataset features a collection of 10073 images obtained from publicly available multimedia repositories.

To evaluate our model's performance, we use Pearson's Linear Correlation Coefficient (PLCC) and Spearman's Rank order Correlation Coefficient (SRCC) as metrics. PLCC measures the model's prediction accuracy, while SRCC evaluates the monotonicity of the BIQA algorithm predictions. Both metrics range from 0 to 1, with higher values indicating better performance in prediction accuracy and monotonicity. We divide the data into 80\% for training and 20\% for testing. To reduce performance bias, we conducted each experiment 10 times and calculated the median PLCC and SRCC.

\subsection{Implementation Details}
\subsubsection{Models}
Our image encoder and text encoder are both sourced from the CLIP framework. Specifically, we employ the ViT-B/16 architecture and a three-layer transformer decoder for the image encoder. The encoder encompasses 12 transformer layers, each comprising a hidden size of 768 dimensions. To match the output of the text encoder, we reduce the dimension of the image feature vector from 768 to 512 using a linear layer.

\subsubsection{Training Details}
Our model is trained for 30 epochs, and 5 epochs for warming up. We employ the Adam optimizer, initialized with a learning rate of $3\times10^{-5}$, which decays using a cosine schedule. During training and inference, we randomly crop 8 sub-images with a spatial size of 224×224×3 from
the original images. We set the coefficients $\lambda_1$ to 1.0, set $\lambda_2$ to 0.1 for LIVE and CSIQ and to 0.5 for other datasets. The batch size is set at 32 for LIVE, CSIQ, and BID datasets, and at 64 for the remaining datasets. 
%

\begin{table}[t]
\caption{Performance comparison is measured by averages of SRCC and PLCC. The best results are highlighted in bold.} 
\setlength\tabcolsep{2pt}
\small
  \centering
  
    \begin{tabular}{lccccc}
    \toprule
     & \multicolumn{2}{c}{LIVEC} & \multicolumn{2}{c}{KonIQ}\\
    \cmidrule{2-5}
    Method & \multicolumn{1}{c}{PLCC} & \multicolumn{1}{c}{SRCC} & \multicolumn{1}{c}{PLCC} & \multicolumn{1}{c}{SRCC} \\
    \midrule
    CLIP-IQA &0.594 & 0.612  & 0.727 & 0.695
    \\
    CLIP-IQA$^{+}$ &0.832 & 0.805  & \textbf{0.909} & 0.895 
    \\
    
    \midrule
    MP-IQE(ours) & \textbf{0.916} & \textbf{0.900} & 0.904 & \textbf{0.898}
    \\
    
    \bottomrule
    \end{tabular}
  
  \label{table2}
\end{table}

\begin{table}[t]
\caption{SRCC on the cross datasets validation. The best
performances are highlighted in boldface.} 
\label{table3}
\setlength\tabcolsep{6pt}
\small
  \centering
    \begin{tabular}{c|cccc}
    \toprule
    Training &LIVEC & KonIQ & LIVE & CSIQ  \\
    \midrule
    Testing & KonIQ & LIVEC &CSIQ & LIVE  \\
    \midrule
    DBCNN & 0.754  & 0.755  & 0.758  & 0.877 \\
    P2P-BM & 0.740  & 0.770 & 0.712  & - \\
    HyperIQA & 0.772  & 0.785  & 0.744  & 0.926 \\
    TReS & 0.733  & 0.786  & 0.761  & - \\
    
    \midrule
    Ours & \textbf{0.781} & \textbf{0.844} & \textbf{0.783} & \textbf{0.935} \\
    \bottomrule
    \end{tabular}%

\end{table}

\begin{table}[t]
    \caption{Ablation study on each component.}
  \label{table4}
\setlength\tabcolsep{4pt}
\small
  \centering
    \begin{tabular}{c|cc|cc}
    \toprule
    \multirow{2}{*}{Variants} & \multicolumn{2}{c|}{CSIQ} & \multicolumn{2}{c}{LIVEC}\\
    \cmidrule{2-5} 
    & PLCC & SRCC & PLCC & SRCC\\
    \midrule
    MP-IQE &\textbf{0.968} & \textbf{0.961} & \textbf{0.916} & \textbf{0.900}\\
    w/o Scene prompts  &0.965 & 0.956 & 0.913 & 0.882\\
    w/o Distortion prompts & 0.961 & 0.953 & 0.911 & 0.884\\
    w/o Deep visual prompts & 0.947 & 0.932 & 0.893 & 0.874 \\
    Shallow visual prompts & 0.956 & 0.947 & 0.905 & 0.888 \\
    \bottomrule
    \end{tabular}%

\end{table}

\subsection{Overall Prediction Performance Comparison}
Table \ref{table1} compares the performance of our MP-IQE with the top 9 state-of-the-art methods, including hand-crafted methods like ILNIQE, CNN-based methods such as DBCNN, and transformer-based methods like MUSIQ. We assess our approach using six common image quality evaluation datasets, covering a variety of images with different distortions and artifacts. Notably, MP-IQE exhibits competitive performance across these datasets. Specifically, on the CSIQ, and KADID datasets, we outperform the current state-of-the-art, achieving significant improvements of 1.5\% and 1.1\% improvements on the SRCC indicator and 0.9\% and 1.3\% on the PLCC indicator.
We attribute these results to multi-modal prompt design and the dual-prompt scheme we employ. Firstly, multi-modal prompt design enables the model to incorporate textual and visual information simultaneously, providing a more holistic understanding of quality-related features in images. Secondly, the dual-prompt scheme guides the model to effectively extract quality-related image features, achieving a comprehensive understanding of image quality. 

Table \ref{table2} compares MP-IQE with CLIP-IQA and CLIP-IQA$^{+}$. Our model outperforms CLIP-IQA and CLIP-IQA$^{+}$ in performance on the LIVEC dataset when compared to other CLIP-based models. However, its advantage narrows considerably when evaluated on the KonIQ dataset. We attribute this to the diverse range of scene content and types of distortions present in the KonIQ dataset. Our approach does not dissect these elements with sufficient granularity, potentially allowing extraneous noise to seep into the analysis. Consequently, the improvement in performance on the KonIQ dataset is not as pronounced.
In summary, these results emphasize the excellent performance of MP-IQE and highlight the potential of the CLIP-based model in advancing Image Quality Assessment.

\subsection{Generalization Capability Validation}

To evaluate the generalization capability of our proposed MP-IQE, we perform rigorous experiments on additional unseen datasets. Specifically, we train the model on one dataset and then test it directly on another unseen dataset. The medians of SRCC for cross-dataset validation on four datasets are reported in Table \ref{table3}. As observed, MP-IQE achieves the best performance on all four datasets. This success can be attributed to several factors.

Firstly, MP-IQE likely benefits from the robust and diverse feature representations that the pre-trained CLIP model encapsulates. The rich visual and textual embeddings inherent in CLIP provide a strong foundation for understanding image quality and content types.

Secondly, our multi-modal prompt learning approach maintains the CLIP model's encoders in a frozen state, which preserves its intrinsic generalization strength. By not fine-tuning the encoders, we avoid overfitting to the peculiarities of the training dataset, allowing the model to focus on the underlying aspects of image quality that are common across datasets, rather than dataset-specific features. This approach encourages the model to develop a more universal understanding of image quality.
%



\subsection{Ablation Study}
\subsubsection{Learnable Text Prompts}
To evaluate the necessity of the dual-prompt scheme in the text branch, we introduce the following variations in our study: 
(1) Without scene prompts: this variation removes the scene prompts from the text branch and keeps the distortion prompts. (2) Without distortion prompts: this variation removes the distortion prompts from the text branch and keeps the scene prompts.

The ablation study results in Table \ref{table4} underscore the essential role of scene prompts and distortion prompts for MP-IQE. 
In the text branch, scene prompts and distortion prompts are responsible for different functions. Scene prompts help the model understand the contextual environment of images, which is crucial for accurately evaluating image quality. When removing scene prompts, the model's ability to capture contextual understanding of image quality is affected, leading to performance degradation. Similarly, distortion prompts enable the model to accurately identify and evaluate specific distortions in the image. If there is no distortion prompt, the professionalism of the model in locating and evaluating image distortion will be compromised. These prompts provide linguistic guidance to the model in the form of text, enabling it to better interpret and analyze image content.
%

\subsubsection{Learnable Deep Visual Prompts}
The role of deep visual prompts in the model's architecture is pivotal for accurate image quality evaluation, as they enable the model to fine-tune its visual representations. These prompts serve as intuitive visual clues, presented in the form of images, which assist the model in better understanding and processing the visual data. However, when we experiment with removing these prompts from the visual branch, we effectively strip the model of this crucial visual guidance. The impact of this removal is evident in the performance metrics, as indicated in Table \ref{table4}. Specifically, the model's performance, as measured by the SRCC metric, drops by 2.9\% on the CSIQ dataset and by 2.6\% on the LIVEC dataset.
Moreover, we introduce shallow visual prompts, incorporating prompts only at the initial layer of the model. In Table \ref{table4}, we observe that the effectiveness is not as pronounced as when prompts are integrated at all layers. This suggests that the presence of visual prompts at deeper levels of the model's architecture is essential for the model to adjust its visual representation capabilities comprehensively, which in turn is critical for the nuanced task of image quality evaluation.

\begin{table}[t]
    \caption{Data-Efficient Learning Validation.}
  \label{table6}
\setlength\tabcolsep{4pt}
\small
\centering
\begin{tabular}{clcccc}
    \toprule
     &  & \multicolumn{2}{c}{LIVE} & \multicolumn{2}{c}{LIVEC}  \\ \cmidrule{3-6} 
    Training Set & Methods & PLCC & SRCC & PLCC & SRCC \\ 
    \midrule
    \multirow{4}{*}{\large 20\%}  & ViT-BIQA & 0.828 & 0.894 & 0.641 & 0.662 \\
     & HyperNet  & 0.950 & 0.951 & 0.809 & 0.776 \\
     & Ours & \textbf{0.960} & \textbf{0.956} & \textbf{0.845} & \textbf{0.808} \\  
     \midrule
     \multirow{4}{*}{\large 40\%} & ViT-BIQA & 0.847 & 0.903 & 0.714 & 0.684  \\
     & HyperNet  & 0.961 & 0.959 & 0.849 & 0.832 \\
     & Ours & \textbf{0.974} & \textbf{0.972} & \textbf{0.889} & \textbf{0.856}\\  
      \midrule
     \multirow{4}{*}{\large 60\%} & ViT-BIQA & 0.856 & 0.915 & 0.739 & 0.705  \\
     & HyperNet  & 0.963 & 0.960 & 0.862 & 0.843 \\
     & Ours & \textbf{0.975}  & \textbf{0.973} & \textbf{0.904 }& \textbf{0.880} \\  
     \bottomrule
\end{tabular}%

\end{table}

\begin{table}[t]
    \caption{Sensitivity study on learnable text prompt length and
visual prompt length. \textcircled{1} denotes text prompt, \textcircled{2} denotes visual prompt.}
  \label{table5}
\setlength\tabcolsep{4pt}
\small
  \centering
    \begin{tabular}{cc|cc|cc}
    \toprule
    \multirow{2}{*}{Prompt} & \multirow{2}{*}{length} & \multicolumn{2}{c|}{LIVE} & \multicolumn{2}{c}{CSIQ}\\
    \cmidrule{3-6} 
    && PLCC & SRCC & PLCC & SRCC\\
    \midrule
    \multirow{4}{*}{\textcircled{1}} & 2 & 0.977 & 0.974 & 0.964 & 0.956 \\
    & 4 & 0.979 & 0.975 & 0.963 & 0.953\\
    & 8 & \textbf{0.980} & \textbf{0.978} & \textbf{0.968} & \textbf{0.961}\\
    & 16 & \textbf{0.980} & \textbf{0.978} & 0.963 & 0.952\\
    \midrule
    \multirow{4}{*}{\textcircled{2}} & 1 & 0.978 & 0.974 & 0.965 & 0.957 \\
    & 2 &0.978 &0.975 & \textbf{0.969} & \textbf{0.961} \\
    & 4 & \textbf{0.980} & \textbf{0.978} & 0.968 & \textbf{0.961}\\
    & 8 & 0.978 & 0.974 & 0.967 & 0.959\\
    \bottomrule 
    \end{tabular}%
    \vspace{-2mm}
\end{table}

\subsection{Data-Efficient Learning Validation.}
Due to the high cost of data annotation for BIQA, the Data-Efficient Learning characteristic of models has gradually become an important metric for evaluating IQA models. 
We assess the use of varying amounts of training data from 20\% to 60\%, with an interval of 20\%, to further investigate this property. This process is repeated ten times for each training data volume, with the median of the performance being reported.
The empirical results obtained from the evaluation of the LIVE and LIVEC datasets are promising. As shown in Table \ref{table6}, our method significantly outperforms other methods under the same training data conditions. 
This superior performance can be attributed to the dual-prompt scheme we introduced helps the model capture features related to image quality, allowing it to learn effective representations even with less data.

The dual-prompt scheme's effectiveness can be attributed to several factors. Firstly, it enhances feature relevance by guiding the model's focus towards the most pertinent aspects of image quality, such as scene context and specific distortions. This targeted approach reduces the cognitive load on the model, allowing it to learn from fewer examples without sacrificing performance. Secondly, the scheme promotes learning efficiency by structuring the learning process around key image quality indicators, which can lead to faster convergence and a more robust understanding of the task at hand.

    
    
    

\begin{figure*}[htbp]
  \centering
  \includegraphics[width=1.0\textwidth]{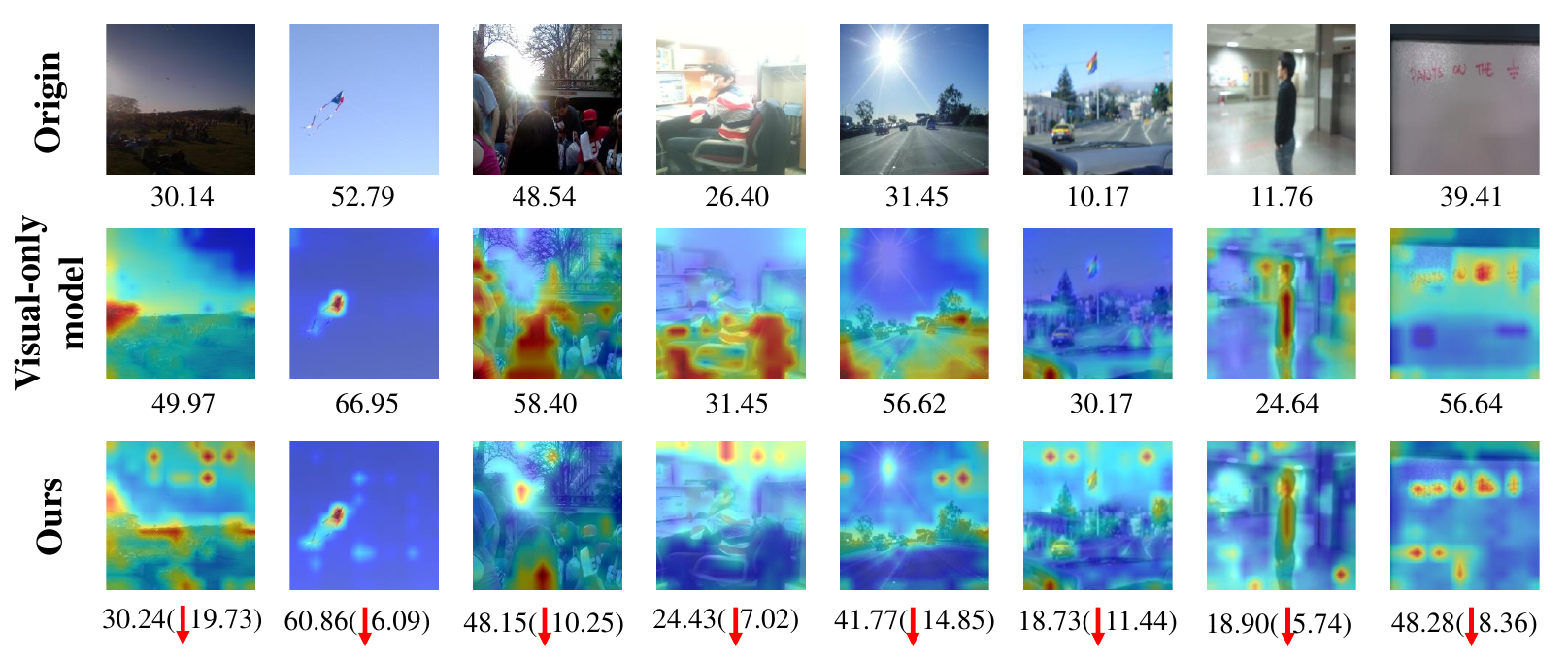}
  \caption{Comparison of activation maps between the visual-only model and our model using Grad-CAM. Rows 1-3 respectively display the original images, CAMs from the visual-only model, and CAMs from our model. The first line of numbers beneath each image represents the ground truth of the original images. The numbers in lines 2-3 represent the predicted quality score, with the numbers in parentheses indicating the distance between our model and the visual-only model.} 
  \label{fig5}
\end{figure*}

\subsection{Sensitivity study of Prompt Length}
We begin our investigation by examining the effects of various prompt lengths on the model's performance on LIVE and CSIQ. As shown in Table \ref{table5}, different prompt lengths have little impact on performance indicators for each dataset. For instance, on the LIVE dataset, the model achieves relatively high performance with visual prompt lengths of 4, while performance slightly decreases with a prompt length of 8. 

From the overall trend, the model generally exhibits better performance as the prompt length increases, although there may be slight decreases in some cases. This suggests that within a certain prompt length range, the model can accurately assess image quality. When selecting prompt length, balancing the model's performance and computational complexity is necessary. Short prompts might lose essential information, while long prompts could introduce unnecessary noise and decrease the training efficiency. Thus, we set the lengths of learnable text prompts and visual prompts to 8 and 4, respectively.

\begin{figure}[htbp]
  \centering
  \includegraphics[width=0.48\textwidth]{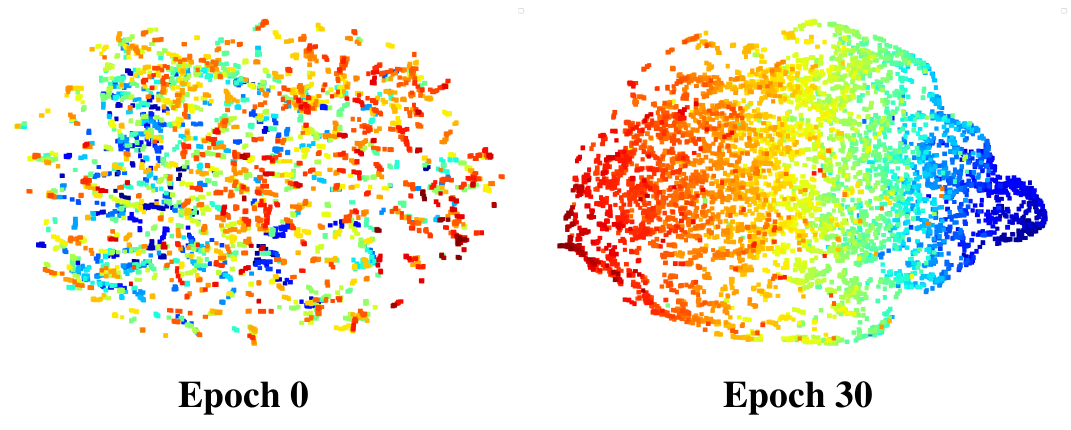}
  \caption{The t-SNE visualization presents the semantic features of the LIVEC training set using our MP-IQE. Different colors in the visualization represent different quality score intervals.} 
  \label{fig6}
  \vspace{-2mm}
\end{figure}

\subsection{Visualization}

\subsubsection{Attention Visualization}
We attempt to explore the importance of introducing text information to the task of image quality assessment. Specifically, we introduce a visual-only model, which includes an encoder with 12 transformer layers and a decoder with a single transformer layer. 
We employ GradCAM~\cite{grad} to illustrate the feature attention maps of the input images in both the visual-only model and our model. Our focus is on low-scoring images, with the aim to reveal the reasons behind our model's advantage on LIVEC. 
As depicted in columns 3, 4, and 5 of Fig.~\ref{fig5}, we noticed that the visual-only model overly focuses on distortion within the main content of images. This results in scenarios where the visual-only model predicts higher quality scores even when the main content of an image is clear, but severe distortions exist in other regions. In contrast, our model takes into account distortions across different areas of an image, leading to a more accurate image quality assessment. 
We attribute this advantage to the dual-prompt scheme. The dual-prompt scheme allows the model to achieve a more comprehensive understanding of the inherent features within images. Especially in low-quality images, our model can effectively identify subtle distortions that might otherwise be overlooked.

As shown in the other columns of the figure, we noticed that our model has clear boundaries in attention to different distorted objects compared to the visual-only model, indicating that our model has better semantic understanding ability. By more fully modeling the object relationships in images, the ability to evaluate image quality can be improved. We attribute this phenomenon to two reasons: firstly, the strong semantic understanding ability of the CLIP, and secondly, the scene prompt we introduced, which helps the model to model the object relationships in the image.

\subsubsection{Semantic Features Visualization}
We employ t-SNE~\cite{van2008visualizing} to illustrate the learned semantic features of the LIVEC dataset using our MP-IQE. 
We use the score interval as the label for t-SNE, for example, the label for an image with a quality score of 45.2 is [45].
As shown in Figure \ref{fig6}, initially, we observe that features of different categories are intertwined and mixed. After training, we can see that our model aggregates features of the same category together, and there is no clear boundary between each category. 
This phenomenon of no clear boundaries aligns with mathematical logic, because in our visualization, even though images with scores of 69.99 and 70.01 belong to categories 69 and 70 respectively, their features should be very similar due to the proximity of their scores.
This result confirms that our model possesses a strong capability for image quality assessment.



\section{Conclusion}
In this study, we explore the potential of CLIP for Blind Image Quality Assessment using the innovative framework MP-IQE. To utilize CLIP's ability to align text and image, we introduce multi-modal prompt learning to improve its comprehension of image content and quality. This innovative strategy effectively overcomes the performance limitations observed in CLIP-IQA and CLIP-IQA$^{+}$. 
In particular, the dual-prompt scheme we introduced aids the model in understanding the IQA task through multi-modal feature alignment, thereby learning more effective quality-related feature representations. Meanwhile, the visual prompts facilitate the model's adaptation to the BIQA domain.
Experimental results demonstrate MP-IQE's superiority over state-of-the-art methods on diverse datasets, affirming its viability for BIQA tasks.

\bibliography{aaai24}
\end{document}